\documentclass[runningheads]{llncs}

\usepackage{eccv}

\usepackage{eccvabbrv}

\usepackage{graphicx}
\usepackage{booktabs}

\usepackage[accsupp]{axessibility}  
\usepackage{multirow}
\usepackage{graphicx}

\usepackage{hyperref}

\usepackage{orcidlink}

\definecolor{tabfirst}{rgb}{1, 0.7, 0.7} 
\definecolor{tabsecond}{rgb}{1, 0.85, 0.7} 
\definecolor{tabthird}{rgb}{1, 1, 0.7} 
\definecolor{lightgray}{rgb}{0.7, 0.7, 0.7}
\usepackage[table]{xcolor}
\usepackage{wrapfig}

\begin{document}

\title{SMG: Semantic Motion Graph for Monocular Dynamic Gaussian Splatting} 

\titlerunning{SMG}


\author{Haozheng Yu \and
Xinyu Yang  \and 
Rundong Luo  \and \\
Jennifer J. Sun
\and Bharath Hariharan}


\authorrunning{Yu et al.}

\institute{Cornell University}

\maketitle


\begin{center}
    \centering
    \captionsetup{type=figure}
    \includegraphics[width=1\textwidth]{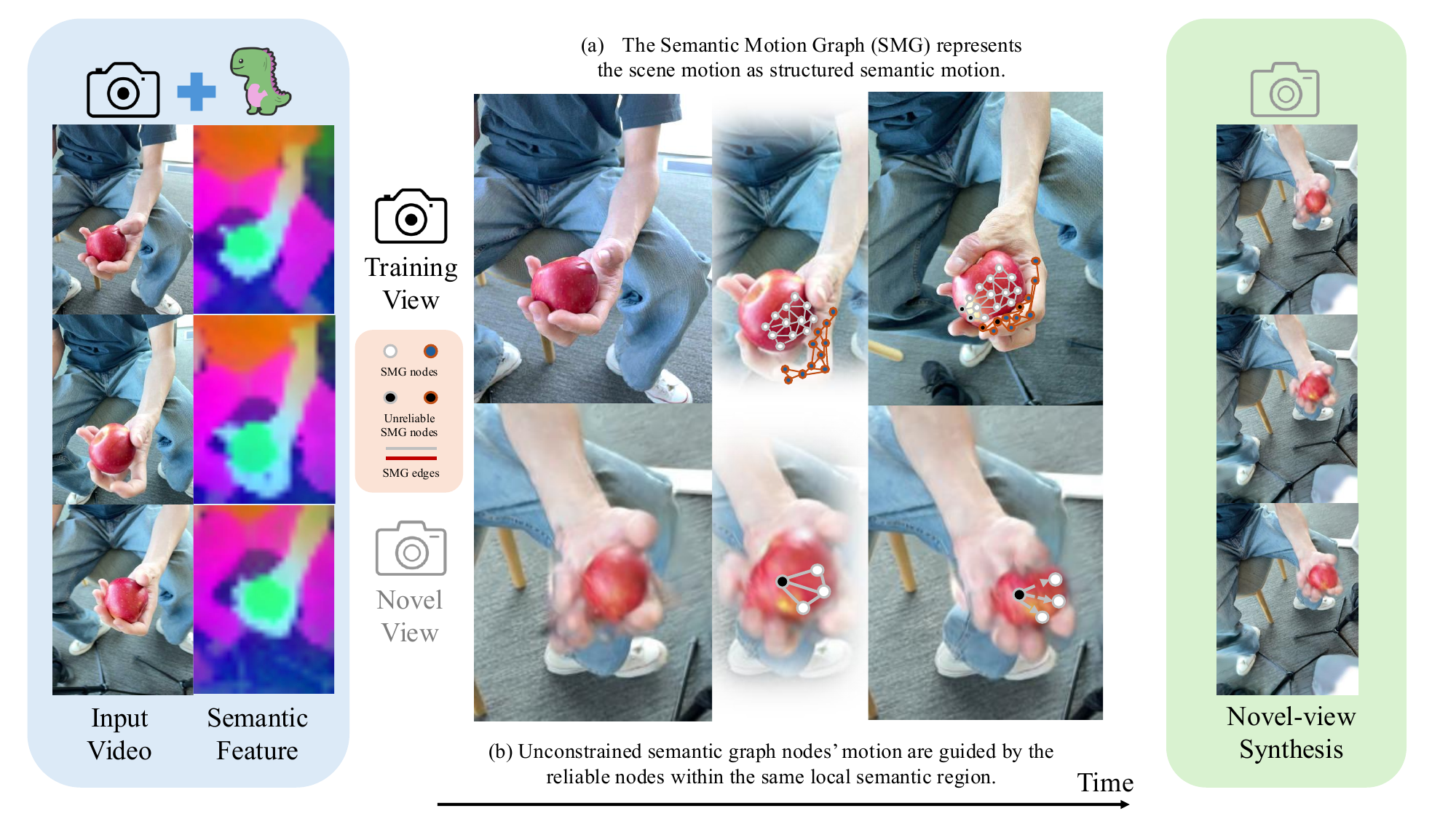}
    \captionof{figure}{We present Semantic Motion Graph (SMG), a monocular dynamic Gaussian splatting framework that models the Gaussian motion as structured semantic motion. SMG handles the uncertainty within the Gaussian optimization. (a). To handle the uncertainty of the Gaussian motion with limited observations, the reliable SMG nodes are used to guide the unreliable motion (b).
    Our method yields robust dynamic Gaussians that generalize to novel views under challenging scenarios. 
    }
    \label{fig:first}
\end{center}%

\begin{abstract}
We study dynamic Gaussian Splatting from monocular videos.
While recent advancements in dynamic Gaussian Splatting offer a promising foundation for modeling dynamic scenes, they often overfit to the training views and fail under occlusion or complex scene motion due to the lack of reliable regularization signals in under-constrained regions.
We propose Semantic Motion Graph (SMG), a novel approach that models the Gaussian motion as the semantic motion.
Our key insight is that the real-world scene motion is often structured by semantic coherence: regions that are spatially close and semantically related tend to exhibit consistent dynamics. 
To leverage this prior, we build SMG to model structured motion of the scene. The Gaussian motion is driven by the motion of SMG nodes.
We further observe that the uncertainty of Gaussian motion arises from both unreliable off-the-shelf priors and weakly constrained regions during optimization. 
SMG addresses this by using reliable graph nodes to guide the motion of nearby unreliable nodes.
To evaluate dynamic Gaussian Splatting under challenging real-world scenarios, we introduce a new multiview dataset collected under an ego-exo setup. 
Extensive experiments demonstrate that SMG achieves state-of-the-art performance on monocular dynamic Gaussian Splatting across challenging real-world benchmarks. Project: \href{https://smg-gaussian.github.io/}{https://smg-gaussian.github.io/}.

\end{abstract}


\section{Introduction}
\label{sec:intro}
Reconstructing the dynamic scene from a monocular video is a crucial task for AR/VR, robotics, and motion capture, as it enables interaction, re-rendering, and 3D control. While recent advances in static 3D scene reconstruction, especially neural-field-based approaches~\cite{mildenhall2021nerf, muller2022instantngp, zhang2020nerf++, barron2021mip, barron2023zip, fridovich2022plenoxels}, have achieved remarkable photorealism, these approaches often break down when extended to dynamic scenes. The reason is fundamental: real-world videos rarely provide dense viewpoints, and object motion is inherently unpredictable. Without sufficient multi-view constraints, distinguishing camera motion from object motion becomes extremely ill-posed. Occlusions and missing observations leave large portions of the scene unsupervised, causing severe overfitting, surface drifting, and inconsistent geometry when viewed from novel perspectives.

As a popular scene reconstruction representation, Gaussian Splatting also suffers under sparse viewpoints. Many Gaussians are only weakly constrained, and their colors and opacities often collude to fit the observed views while hallucinating geometry in unobserved regions, leading to floating artifacts and inconsistent reconstruction~\cite{li2024dngaussian, zhu2024fsgs, zhang2024corgs, chen2025quantifying}. Recent efforts~\cite{liu2025modgs, wang2025som, lei2025mosca, stearns2024dynamicmarbles} introduce data-driven priors and physically-based priors~\cite{luiten2023dynamic} to regularize the dynamic Gaussian optimization under limited viewpoints. 
These methods assume that the irregular and complex per-frame motion can be decomposed into smooth and local rigid transformations, which regularizes dynamic Gaussian optimization under limited views. Yet, motion smoothness alone is insufficient: in many real scenes, occlusion hides entire body parts or objects, and purely geometric constraints fail to propagate correct motion into such regions.

In this work, we take the idea one step further. 
We observe that local rigidity is not merely a geometric assumption—it emerges from semantic consistency. Gaussians belonging to the same semantic region (e.g., a rigid object) tend to share coherent motion. 
Motivated by this, we introduce Semantic Motion Graph (SMG), a new mechanism that models the motion of Gaussians as SMG node deformation (Fig.~\ref{fig:first}).
We first build the SMG based on lifted semantic tracks, which restrict motion propagation within local semantic groups. 
We also observe that the uncertainty of Gaussian motion arises from both unreliable priors and unconstrained regions during optimization. 
To address this issue, we propose confidence-aware as-rigid-as-possible (C-ARAP), a mechanism that propagates motion and maintains topology by guiding less reliable nodes with reliable ones. 
We further design local rigid motion control (LRM), which handles uncontrollable node velocity changes when observations are limited. 
Together, these mechanisms provide additional regularization for SMG, enabling correct motion propagation and better generalization to challenging novel views.

To evaluate dynamic Gaussian Splatting under complex camera and object motion, we propose SMG dataset, a custom multiview dataset captured in an ego-exo setup. Together with SMG, they enable a more comprehensive understanding of dynamic Gaussian Splatting under challenging real-world conditions.

To summarize, our contributions are: 
\begin{itemize}
\item We propose SMG, a monocular dynamic Gaussian Splatting pipeline that models the Gaussian motion as a semantic motion graph. 
\item We develop SMG-based motion uncertainty modeling that propagates motion from reliable nodes to less visible nodes within the same semantic group, thereby handling uncertainty during dynamic Gaussian optimization.
\item We introduce SMG dataset, a challenging multiview dataset with complex camera and object motion for monocular dynamic Gaussian Splatting.
\item SMG demonstrates state-of-the-art performance for novel-view synthesis on challenging scenes and can also benefit other scene understanding and 3D tasks, e.g. 3D tracking.

\end{itemize}
\section{Related Work}
\label{sec:related}


\subsection{Dynamic Scene Representation}

Reconstructing and synthesizing the real world is a fundamental question in the computer vision community. NeRF~\cite{mildenhall2021nerf} is one of the pioneering works in this field, which introduced the idea of learning a neural radiance field implicitly with a multi-layer perceptron (MLP). This approach also enables novel-view synthesis via volumetric rendering. However, optimization and rendering via the MLP is time-consuming, even when using GPUs. A series of follow-up works address the rendering speed~\cite{chen2022tensorf, fridovich2022plenoxels, muller2022instantngp} or the rendering quality~\cite{zhang2020nerf++, barron2021mip, barron2022mip360, barron2023zip} issues of NeRF. Some works further explored using NeRF to represent complex dynamic scenes, for example, by modeling a deformation field based on a canonical field~\cite{du2021neural, pumarola2021dnerf, park2021nerfies,fang2022tineuvox}, directly learning a neural radiance field conditioned on time~\cite{gao2021dynamic, li2022neural, park2021hypernerf}, or learning a hybrid representation to parameterize the dynamic scene~\cite{cao2023hexplane, fridovich2023kplane}. 

Recently, 3D Gaussian Splatting (3DGS)~\cite{kerbl20233dgs} has gained significant attention for its efficient rendering, fast training speed, and outstanding visual effects. Unlike NeRF, 3DGS represents the scene as a set of 3D Gaussians without the MLP. 
Built on the success of 3DGS~\cite{kerbl20233dgs}, following works have further expanded this representation to model dynamic scenes by encoding the 3D Gaussian position alongside its translation and rotation over time~\cite{luiten2023dynamic}, optimizing the deformation field of the canonical Gaussians~\cite{yang2024deformable, Wu_2024_4dgs}, or modeling time as an additional dimension~\cite{yang2023gs4d}. However, when applied to real-world videos with sparse viewpoints, 3D Gaussians often overfit the training views and produce unreliable reconstructions from novel views~\cite{li2024dngaussian, zhu2024fsgs, zhang2024corgs, chen2025quantifying}. 
Recent and concurrent efforts explore learning feed-forward transformer-based models to reconstruct dynamic Gaussians~\cite{xu20254dgt}, or incorporating strong data-driven priors to regularize the optimization of dynamic Gaussians and enable reliable novel view synthesis from in-the-wild videos~\cite{liu2025modgs, stearns2024dynamicmarbles, seidenschwarz2025dynomo, lei2025mosca, park2025splinegs, wang2025som, wu2025origs}. 
However, under severe occlusions or when the learned priors do not generalize, these methods may still fail. We observe that real-world motion is semantically structured, and introduce a semantic-aware Gaussian motion propagation framework that regularizes the deformation of under-constrained Gaussians, leading to more stable dynamics and improved novel-view synthesis.

\subsection{3D Feature Field}
With the development of vision foundation models~\cite{radford2021clip, li2022lseg, oquab2023dinov2, kirillov2023sam, ravi2024sam2}, recent approaches explore to distill the high-level features into neural fields. Early effort~\cite{zhi2021place} encodes semantic labels in NeRF~\cite{mildenhall2021nerf} to enable semantic completion from noisy 2D labels. LERF~\cite{kerr2023lerf} and its following works~\cite{liu2023weakly, kim2024garfield} distill CLIP~\cite{radford2021clip} and DINO~\cite{caron2021dino} features into NeRF to enable open-vocabulary 3D queries. Recent approaches~\cite{cen2023saga, qin2024langsplat, zhou2024feature} focus on distilling the 3D feature fields with Gaussian splatting, enabling efficient 3D segmentation and understanding.


Recent works explore distilling semantic fields on top of dynamic Gaussians to enable semantic-driven tracking and Gaussian editing~\cite{labe2024dgd, fiebelman20254legs, song2025coda}. Concurrent efforts~\cite{4dlangsplat, feature4x} further incorporate MLLMs for user-interactive 4D VQA tasks. 
However, these works solely treat semantic fields as an auxiliary output of the dynamic Gaussian Splatting pipelines, ignoring the fact that the dynamics in real-world monocular videos are often semantically structured. 
In contrast, our goal is fundamentally different: we exploit semantic cues as a strong regularization of the Gaussian deformation under sparse observations.
We leverage the semantically structured nature of motion and introduce a semantic motion graph that transforms the Gaussian motion into the semantic graph motion. We further propose uncertainty modeling strategies that guide the graph motion under noisy priors or unreliable optimization, leading to more stable motion propagation and improved novel-view synthesis.

\subsection{3D Scene Understanding}
Our work is also related to the area of 3D scene understanding more broadly. This area includes tasks such as sparse and dense 3D grounding~\cite{achlioptas2020referit3d,chen2020scanrefer,chen2022language,huang2022multi,yang2021sat,zheng2024video}, 3D question answering~\cite{chen2021scan2cap,ma2022sqa3d} and 3D captioning~\cite{chen2021scan2cap,chen2023end}. 3D understanding fundamentally differs from 2D understanding by demanding a comprehensive grasp of spatial geometry, occlusion, and physical properties. However, the scarcity of 3D data and the inherent limitations of 2D data to recover 3D information pose significant challenges for 2D-trained vision foundation models in effectively understanding 3D scenes. Initialized by the 2D priors and the semantics, our work addresses this gap by learning the 3D feature fields and tracks in the process of optimizing dynamic Gaussians.
Our approach, similar to other works based on 3D Gaussians~\cite{4dlangsplat, feature4x}, employs test-time optimization on individual scenes. 
We aim to bridge the dimensional gap and facilitate robust spatial understanding from monocular videos (e.g. 3D object segmentation and tracking), thereby contributing to 3D scene understanding more broadly.

\begin{figure*}[!t]
  \centering
    \includegraphics[width=1\linewidth]{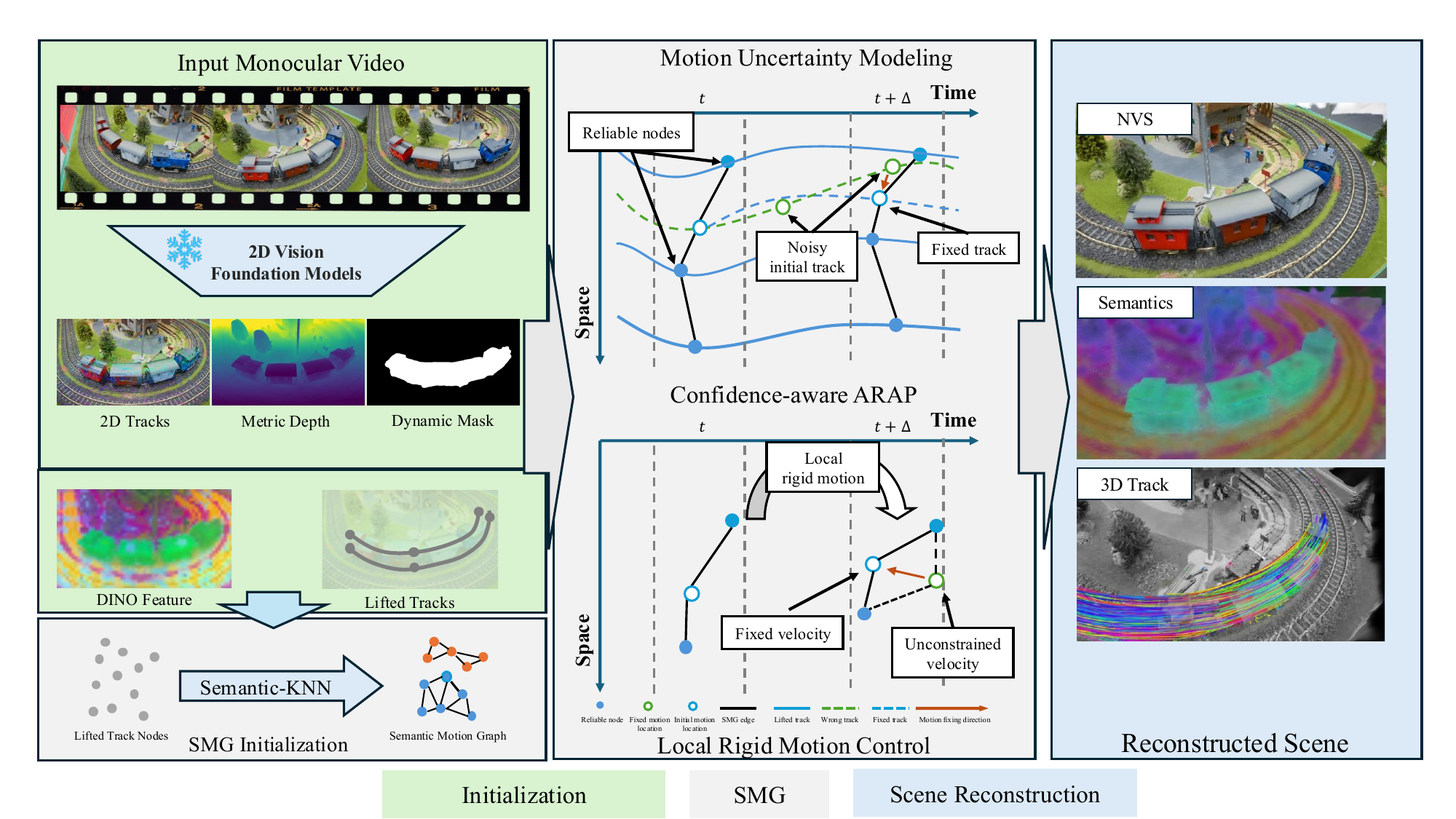}

  {
    \caption{\textbf{Overview of SMG}. Given a monocular video with 2D tracks, depth, and dynamic/static masks predicted by the off-the-shelf models, the Semantic Motion Graph is initialized with the 3D lifting field in the nearby semantic groups. 
    To handle the uncertainty during the Gaussian optimization, we design confidence-aware ARAP (C-ARAP) and local rigid motion control (LRM) to preserve the graph topology and guide the motion of unreliable nodes by the reliable nodes, producing semantically structured and temporally coherent motion. 
    Finally, our proposed SMG improves the robustness of novel-view synthesis and yields dynamic 3D scenes enriched with consistent semantics and 3D tracks.}
    \label{fig:system}
  } 
\end{figure*}

\section{Method}
\label{sec:method}




We introduce Semantic Motion Graph (SMG), a novel method that models the motion of semantic groups for dynamic Gaussian Splatting under a monocular setup. The SMG is first initialized with the 3D lifting field within nearby semantic groups, which ensures that the motion is only propagated locally (Sec.~\ref{sec:semantic mosca}). To handle the uncertainty during the optimization, the confidence-aware as-rigid-as-possible (C-ARAP) and local-rigid-motion (LRM) controls are incorporated into SMG to ensure that the semantic motion is propagated correctly under unreliable supervision (Sec.~\ref{sec: anchors}). Finally, the dynamic Gaussians are initialized and optimized under the guidance of SMG (Sec.~\ref{sec:gaussian_optimization}), yielding dynamic Gaussians that remain stable under novel views.

\subsection{Semantic Motion Graph (SMG)}
\label{sec:semantic mosca}

A major challenge in monocular dynamic Gaussian Splatting is that motion in unobserved regions is inherently ill-posed. Recent methods~\cite{stearns2024dynamicmarbles, wang2025som, lei2025mosca} utilize the strong off-the-shelf priors to construct 3D lifting fields to regularize the motion in the unconstrained regions. 

We argue that resolving such ambiguity requires modeling how motion correlations are structured. In real-world scenarios, the motion coherence is structured by semantic coherence rather than mere geometric proximity. To this end, we present Semantic Motion Graph (SMG) (Fig.~\ref{fig:system}). Given the 2D tracks, visibility, and the depth estimated by the off-the-shelf models, we construct the graph $\mathcal{G} = (\mathcal{V}, \mathcal{E})$ with the lifted tracks as nodes. We restrict connectivity to spatial local neighbors to preserve geometric continuity, 
while enforcing semantic consistency to avoid propagating motion across unrelated nearby objects. Formally, the edge $\mathcal{E}_{ij}$ between nodes $i,j \in \mathcal{V}$ is defined as

\begin{equation}
\mathcal{E}_{ij}
=
\mathbf{1}\!\left[j \in \mathrm{KNN}(i)\right]
\cdot
\mathbf{1}\!\left[\cos(f_i,f_j) \ge \tau\right]
\end{equation}
where $\mathrm{KNN}(\cdot)$ is performed at the trajectory level, using the top-$k$ distances computed over frames in which both $i$ and $j$ are visible. $\tau$ is the semantic gating threshold. $\cos(f_i,f_j)$ denotes the cosine similarity of the semantic feature of $i$ and $j$. To obtain the semantic feature of each trajectory, we first extract a per-frame semantic feature map and project each trajectory point to image coordinates in its visible frames. We then aggregate these frame-wise features by top-k cosine-consistent averaging to obtain a robust trajectory-level semantic descriptor. We use DINOv3~\cite{simeoni2025dinov3} as our semantic feature extractor $\mathrm{f}(\cdot)$ in all our experiments.

\subsection{Modeling Motion Uncertainty in Dynamic Gaussians}
\label{sec: anchors}

Dynamic Gaussian Splatting under monocular setups is inherently ill-posed and prone to uncertainty. 
Noisy off-the-shelf priors can be a source of uncertainty. During optimization, regions with limited visibility are not well constrained, which introduces additional uncertainty into the system.
To address these challenges, we introduce confidence-aware as-rigid-as-possible (C-ARAP) and local rigid motion control (LRM) to handle the uncertainty within the whole optimization process.

\noindent \textbf{Confidence-aware as-rigid-as-possible.} Dynamic Gaussian Splatting under the monocular setup suffers from uncertainty introduced by off-the-shelf priors. Due to occlusions and complex object motions in monocular videos, the estimated depth and tracks inevitably introduce noise to the system. The proposed SMG is initialized from the lifted 2D tracks, which may lead to both reliable and unreliable motion being propagated in the graph. 

We use confidence-aware as-rigid-as-possible (C-ARAP) to propagate motion within the semantic groups. Rigidity modeling of dynamic Gaussians~\cite{luiten2023dynamic} and lifting field nodes~\cite{lei2025mosca} has been introduced by previous works. However, modeling all the Gaussians or nodes fairly introduces uncertainty during motion propagation. We modulate each SMG edge using node-level confidence estimated from visibility and motion consistency. This allows local motion to be propagated by reliable nodes within a local semantic group while suppressing the nodes with high uncertainty to propagate the unreliable motions. Given the per-frame node visibility $\mathrm{viz}_{t,i} \in \{0,1\}$ from the off-the-shelf point tracking model, the per-frame node confidence of node $i$ is defined as
\begin{equation}
c_{t,i} = \mathrm{viz}_{t,i} \cdot (c_{t,i}^{\text{motion}})^{\lambda_m} \cdot (c_{t,i}^{\text{rigid}})^{\lambda_r},
\end{equation}
\begin{equation}
c_{t,i}^{\text{motion}}
=
\exp\!\left(
-
\left(
\frac{r_{t,i}^{v}}
{\kappa_m (s_{t,i}^{v} + \varepsilon)}
\right)^2
\right),
\quad
c_{t,i}^{\text{rigid}}
=
\exp\!\left(
-
\left(
\frac{r_{t,i}^{\ell}}
{\kappa_r (s_{t,i}^{\ell} + \varepsilon)}
\right)^2
\right),
\end{equation}
where $c_{t,i}^{\text{motion}}$ and $c_{t,i}^{\text{rigid}}$ are the framewise node motion and rigid confidence. $r^v$ and $r^l$ denote the deviation of the node velocity from the neighborhood average velocity and the abnormal change of edge length across adjacent frames, respectively; larger values indicate lower reliability due to motion inconsistency or local non-rigidity. $s^v$ and $s^l$ are the absolute deviation over neighbors used for adaptive normalization. $\kappa_m$ and $\kappa_r$ are the soft thresholds controlling the tolerance level. This framewise node confidence determines the reliability of the node motion that reflects the local motion. Once the per-frame node confidence is obtained, we define the asymmetric confidence-aware edge weight of SMG as
\begin{equation}
\tilde{w}_{ij}^{\,t}
=
w_{ij}^{\text{topo}}\left(\alpha + (1-\alpha)c_{t,j}\right),
\end{equation}
where $\tilde{w}_{ij}^{\,t}$ denotes the asymmetric edge weight from node $i$ to node $j$ at time $t$ and $\alpha \in [0,1]$ is a lower bound that prevents the edge weight from collapsing to zero. $w_{ij}^{\text{topo}}$ is the weight of SMG edge $E_{ij}$ normalized by topological connectivity. The motion of nodes with low framewise confidence should be guided by nearby reliable nodes rather than vice versa. Finally, the C-ARAP is computed as
\begin{equation}
L_{\text{C-ARAP}} =
\sum_{t,i} \sum_{j \in \mathcal{N}(i)}
\tilde{w}_{ij}^{\,t}
\left[
\lambda_{c}\,\!\left(\left\| d_{ij}^{\,t+\delta} - d_{ij}^{\,t} \right\|\right)^{2}
+
\lambda_{l}\,\!\left(\left| \| d_{ij}^{\,t+\delta} \| - \| d_{ij}^{\,t} \| \right|\right)^{2}
\right],
\end{equation}
\begin{equation}
d_{ij}^{\,t} = (R_i^{\,t})^{\top}(x_j^{\,t} - x_i^{\,t}),
\end{equation}
where $j \in \mathcal{N}(i)$ is the neighboring nodes of node $i$. $\lambda_c$ and $\lambda_l$ are the parameters that control the relative strength of the vector-consistency term and the length-preservation term. $d_{ij}^{\,t}$ is computed as the multiplication of the orientation $R_i^{\,t}$ of node $i$ at time $t$ and the world displacement $x_j^{\,t} - x_i^{\,t}$ from node $i$ to node $j$ at time $t$. 

\noindent \textbf{Local rigid motion under unconstrained regions.} Beyond the uncertainty introduced by off-the-shelf priors, monocular dynamic Gaussians are also sensitive to the estimated velocity field. When geometric constraints are weak, optimization can produce locally inconsistent velocities, leading to structural drifts.

To stabilize local motion, we introduce a local rigid motion control that constrains the local velocity field to be explained by a rigid twist. Given a node $i$ and its neighboring nodes $j \in \mathcal{N}(i)$, the average velocity of the local region at time $t$, $\bar{v}_{i}^{t}$ is defined as the average velocity of $j$ weighted by the edge weight. The local center ${g}_{i}^{t}$ is defined as the weighted average location of $j$. We compute the residual of the motion of each $j$ as
\begin{equation}
\tilde{v}_{ij}^{\,t} = v_{j}^{\,t} - \bar{v}_{i}^{\,t},
\quad
\tilde{x}_{ij}^{\,t} = x_{j}^{\,t} - g_{i}^{\,t}.
\end{equation}
We then estimate the local angular velocity $\omega_i^{t}$ that best represents local motion via weighted least squares:
\begin{equation}
\omega_i^{t}
=
\arg\min_{\omega}
\sum_{j\in\mathcal{N}(i)}
\tilde{w}_{ij}^{\,t}\,
\left\|
\tilde{v}_{ij}^{\,t}
-
\omega \times \tilde{x}_{ij}^{\,t}
\right\|^2.
\end{equation}
Finally, we penalize the local node velocities that cannot be explained by the estimated local rigid motion. The local rigid motion loss (LRM) is computed as
\begin{equation}
L_{\text{LRM}} =
\sum_{t,i} \sum_{j \in \mathcal{N}(i)}
\tilde{w}_{ij}^{\,t} \rho \left(
\left\|
\tilde{v}_{ij}^{\,t}
-
\omega_i^{t} \times \tilde{x}_{ij}^{\,t}
\right\|_2 \right).
\end{equation}
where $\rho(\cdot)$ denotes the Huber robust penalty.

\subsection{SMG-guided Dynamic Gaussian Optimization}\label{sec:gaussian_optimization}

\noindent \textbf{Gaussian initialization and deformation.}
After obtaining the SMG, we optimize dynamic Gaussians by coupling photometric supervision with the SMG-based semantic motion regularizations mentioned above.

Given the dynamic mask computed by an off-the-shelf flow model, we sample pixels from dynamic regions with valid depth and back-project them to 3D to initialize dynamic Gaussians.
Each Gaussian is parameterized as
$
\mathcal{G}_i=\{\mathbf{x}_i^{ref},\mathbf{R}_i^{ref},\mathbf{s}_i,\alpha_i,\mathbf{f}_i,t_i^{ref}\}$, 
where $\mathbf{x}_i^{ref}\in\mathbb{R}^3$ is the reference position, $\mathbf{R}_i^{ref}\in SO(3)$ is the reference rotation, $\mathbf{s}_i\in\mathbb{R}^3$ is the scale, $\alpha_i\in(0,1)$ is opacity, $\mathbf{f}_i$ denotes SH coefficients and $t_i^{ref}$ is the reference timestamp.

Each Gaussian is attached to its nearest SMG node at $t_i^{ref}$ as a local anchor.
Instead of being controlled by a single node, it is driven by $K$ semantic neighbors of that node with distance-aware normalized weights $\{w_{ik}\}_{k=1}^K$.
At timestep $t$, we obtain the Gaussian pose by blending node motions from $t_i^{ref}$ to $t$:
\begin{equation}
(\mathbf{x}_i^{t}, \mathbf{R}_i^{t})
=
\mathcal{W}\!\left(
\{ \mathcal{T}_{k}^{\,t_i^{ref}\rightarrow t} \}_{k=1}^{K},
\{ w_{ik} \}_{k=1}^{K},
\mathbf{x}_i^{ref}, \mathbf{R}_i^{ref}
\right), \quad
\sum_{k=1}^{K} w_{ik}=1.
\end{equation}
where $\mathcal{T}_{k}^{\,t_i^{ref}\rightarrow t}$ is the rigid motion of the $k$-th controlling SMG node from reference time to target time, and $\mathcal{W}(\cdot)$ is the weighted motion blending operator.
We follow MoSca~\cite{lei2025mosca} and implement $\mathcal{W}(\cdot)$ with dual-quaternion blending (DQB)~\cite{kavan2007dqb} for stable and smooth articulated deformation.

\noindent \textbf{Photometric Joint Optimization.}
We jointly optimize photometric rendering consistency and semantic motion consistency with the ground truth image and off-the-shelf priors.
The training objective is:
\begin{equation}
\mathcal{L}
=
\lambda_{rgb}\mathcal{L}_{rgb}
+\lambda_{dep}\mathcal{L}_{dep}
+\lambda_{mask}\mathcal{L}_{mask}
+\lambda_{track}\mathcal{L}_{track}.
\end{equation}

During optimization, we update Gaussian parameters together with the SMG deformation. We further apply density controls to adapt the Gaussian population over time and improve reconstruction quality and stability.

\section{Experiments}
\label{sec:experiments}





\begin{table}[t]
\centering
\small
\resizebox{\linewidth}{!}{
\begin{tabular}{lccc|lccc}
\toprule
~Method~ & mPSNR $\uparrow$ & mSSIM $\uparrow$ & mLPIPS $\downarrow$
& ~Method~ & mPSNR $\uparrow$ & mSSIM $\uparrow$ & mLPIPS $\downarrow$ \\
\midrule
T-NeRF~\cite{gao2022dycheck} & 16.96 & 0.577 & 0.379
& Gauss.Marbles~\cite{stearns2024dynamicmarbles} & 16.72 & - & 0.413 \\

NSFF~\cite{li2021nsff} & 15.46 & 0.551 & 0.396
& DyBluRF~\cite{sun2024dyblurf} & 17.37 & 0.591 & 0.373 \\

Nerfies~\cite{park2021nerfies} & 16.45 & 0.570 & 0.339
& CTNeRF~\cite{miao2024ctnerf} & 17.69 & 0.531 & - \\

HyperNeRF~\cite{park2021hypernerf} & 16.81 & 0.569 & 0.332
& D-NPC~\cite{kappel2025dnpc} & 16.41 & 0.582 & 0.319 \\

PGDVS~\cite{zhao2023pgdvs} & 15.88 & 0.548 & 0.340
& MoSca~\cite{lei2025mosca} & \cellcolor{tabthird}19.32 & \cellcolor{tabsecond}0.706 & \cellcolor{tabsecond}0.264 \\

DyPoint~\cite{zhou2023dypoint} & 16.89 & 0.573 & -
& Shape-of-Motion~\cite{wang2025som}~ & 17.32 & 0.598 & 0.296 \\

DpDy~\cite{wang2024dpdy} & - & - & 0.516
& OriGS*~\cite{wu2025origs} & \cellcolor{lightgray}19.69 & \cellcolor{lightgray}0.716 & \cellcolor{lightgray}0.256 \\

Dyn.Gauss.~\cite{luiten2023dynamic} & 7.29 & - & 0.692
& OriGS~\cite{wu2025origs} & \cellcolor{tabsecond}19.43 & \cellcolor{tabthird}0.695 & \cellcolor{tabthird}0.281 \\

4DGS~\cite{Wu_2024_4dgs} & 13.64 & - & 0.428
& \textbf{Ours} & \cellcolor{tabfirst}19.54 & \cellcolor{tabfirst}0.718 & \cellcolor{tabfirst}0.250 \\
\bottomrule
\end{tabular}
}
\vspace{1mm}
\caption{Comparison of novel view synthesis on the Dycheck (iPhone) dataset~\cite{gao2022dycheck}. Following previous works~\cite{lei2025mosca}, results are reported on half-resolution images. The best, second-best, and third-best results are highlighted in red, orange, and yellow. "*" indicates results reported in the OriGS paper~\cite{wu2025origs}.}
\label{tab:novel}
\end{table}

\subsection{Implementation Details}
Our model is implemented using PyTorch and trained on a single NVIDIA H100 GPU. We perform KNN search based on the 3D trajectory distance, which is aggregated over the three co-visible frames with the smallest distances. The scaffold topology is then constructed using the top-16 nearest neighbors. For C-ARAP, we use $\lambda_m = \lambda_r = 1.0$, $\kappa_m  = \kappa_r = 2.5$, $\alpha = 0.6 $ for all experiments. For the semantic gating parameter $\tau$, we use $\tau = 0.75$ for Dycheck and NVIDIA experiments and $\tau = 0.85$ for SMG-Dataset experiments.
For scene reconstruction experiments, we assessed the quality of our reconstructions using standard metrics: Peak Signal-to-Noise Ratio (PSNR), which quantifies pixel-level similarity; Learned Perceptual Image Patch Similarity (LPIPS)~\cite{zhang2018unreasonable}, which measures perceptual similarity; and Structural Similarity Index (SSIM)~\cite{wang2004image}, which evaluates structural similarity.



\begin{figure}[t]
  \centering
  \includegraphics[width=\linewidth, trim=8mm 8mm 5mm 0]{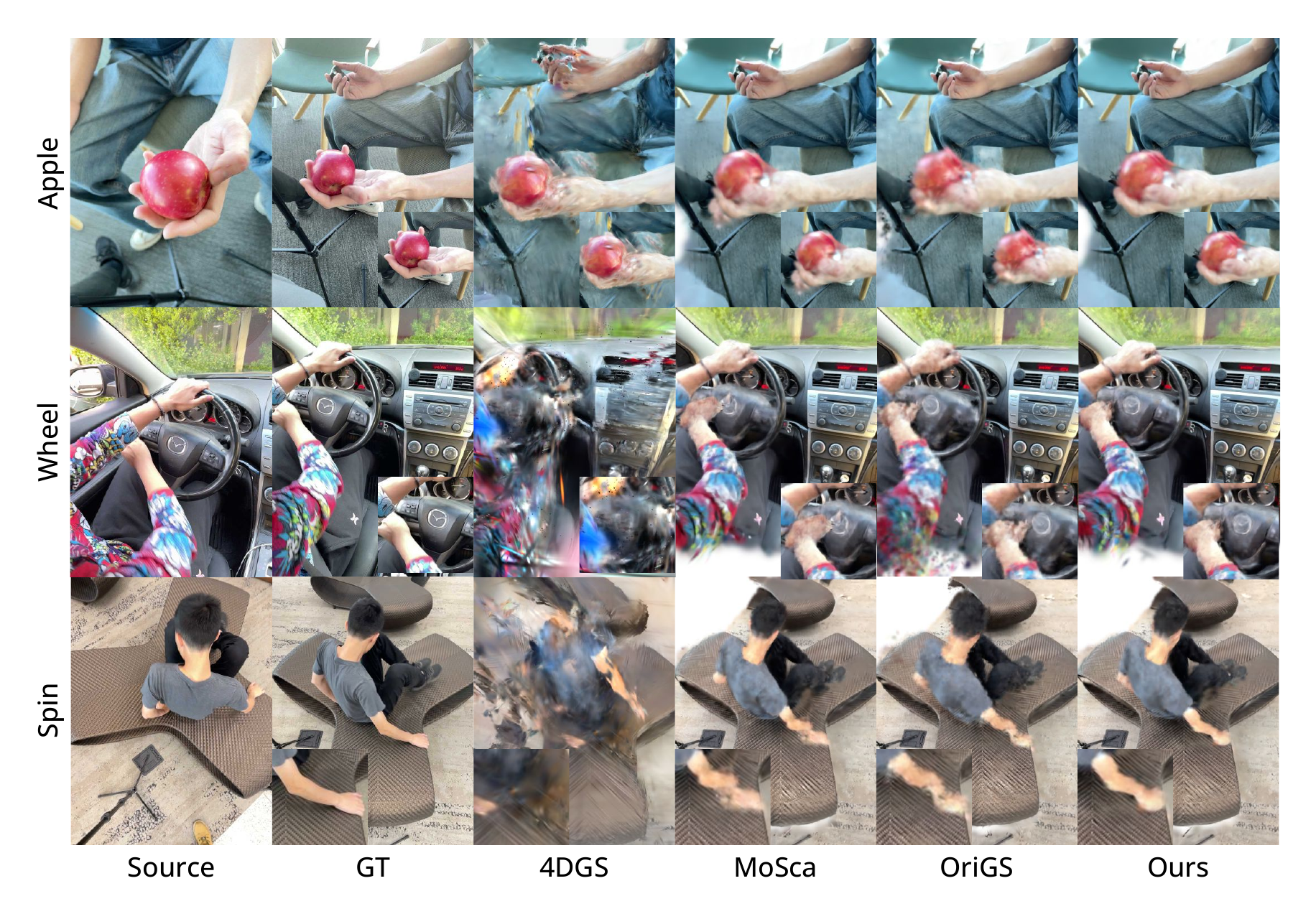}
  \caption{\textbf{Qualitative results of novel-view synthesis on Dycheck dataset.} SMG shows superior results against the baselines. SMG preserves the correct object geometry while effectively reducing floating or drifting dynamic Gaussians, leading to more stable and coherent reconstructions across viewpoints. Zoom-in to see details.}
  \vspace{-5mm}
  \label{fig:semantic}
\end{figure}

\subsection{Evaluations on Dycheck Dataset}
We report our results on Dycheck for novel-view synthesis. Dycheck is an iPhone-captured dataset featuring camera pose and sensor depth. We follow the baselines~\cite{lei2025mosca, stearns2024dynamicmarbles} and evaluate on all 7 scenes: Apple, Block, Paper Windmill, Space Out, Spin, Teddy, and Wheel with half resolution. We use camera view 0 for training and 1 and 2 for testing. Following the previous works, we compute the metrics on the covisible mask area. We report the quantitative results in Table~\ref{tab:novel}. OriGS~\cite{wu2025origs} does not provide the training hyperparameters on the iPhone dataset and the evaluation code. As their experiment follows MoSca's structure, we modify MoSca's hyperparameters and report the reproduced OriGS results. We also provide OriGS paper results as "OriGS*". Due to the wide baseline in some of the Dycheck scenes, e.g. Apple, previous methods that rely solely on photometric optimization fail under novel views. By incorporating off-the-shelf priors for regularization, 3D lifting fields~\cite{lei2025mosca, wang2025som, wu2025origs} achieve much better reconstruction than their counterparts. However, they fail to handle the uncertainty introduced by the monocular priors and the optimization process. For fair comparison, we apply the same off-the-shelf priors with MoSca and OriGS and use the noisy iPhone sensor depth for training. By constructing the semantic lifting field and modeling the uncertainty within the whole optimization process, SMG outperforms the baselines on all metrics. 

Qualitatively, SMG outperforms all baselines on producing less geometry drifting and more reliable object motion (Fig.~\ref{fig:semantic}). The Gaussian deformation field~\cite{Wu_2024_4dgs} produces severe artifacts and fails to reconstruct the object geometry because the deformation MLP tends to overfit to the training views when lacking regularization.  
MoSca~\cite{lei2025mosca} renders overall reasonable scene geometry and motion under novel views when the monocular off-the-shelf priors are reliable. 
However, when the priors are noisy, MoSca mistakenly propagates the motion to the scaffold, leading to the breaking down of the object geometry (the human arm in the "Wheel" and "Spin" scene) and floating Gaussians (e.g. the human hand in the "Apple" scene). OriGS~\cite{wu2025origs} inherits the weaknesses of MoSca while introducing more artifacts and worse geometry.
Compared to the baseline methods, the semantic motion graph ensures that the motion propagation is restricted to nearby semantic regions. 
When the priors are not reliable, the proposed C-ARAP and LRM preserve the object geometry and motion optimized when the supervision is reliable, and reduce the negative effects of the unreliable priors to the motion graph.  
We observe that SMG is good at modeling the dynamic objects under the scenes with a large baseline, e.g. "Apple". SMG better preserves the objects’ physical shapes. This also comes from the confidence-aware uncertainty modeling. Reliable SMG nodes provide stable guidance for unreliable nodes, helping preserve object shape through occlusions and improving temporal consistency. 

\begin{table}
\centering
\vspace{-3mm}
\small
\setlength{\tabcolsep}{3pt}
\resizebox{0.75\linewidth}{!}{
\begin{tabular}{lcc|lcc}
\toprule
Method & PSNR & LPIPS & Method & PSNR & LPIPS \\
\midrule
D-NeRF~\cite{pumarola2021dnerf}         & 21.49 & 0.232 & CTNeRF~\cite{miao2024ctnerf}           & 26.13 & 0.082 \\
NR-NeRF~\cite{tretschk2021nrnerf}       & 19.69 & 0.323 & DynPoint~\cite{zhou2023dypoint}        & \cellcolor{tabthird}26.53 & \cellcolor{tabsecond}0.068 \\
TiNeuVox~\cite{fang2022tineuvox}        & 19.74 & 0.285 & D-NPC~\cite{kappel2025dnpc}            & 25.64 & 0.109 \\
HyperNeRF~\cite{park2021hypernerf}      & 17.60 & 0.367 & RoDynRF~\cite{liu2023rodynrf}          & 25.89 & \cellcolor{tabfirst}0.067 \\
NSFF~\cite{li2021nsff}                  & 24.33 & 0.199 & RoDynRF~\cite{liu2023rodynrf} w/o      & 25.38 & 0.079 \\
DynNeRF~\cite{gao2021dynamic}           & 26.10 & 0.082 & GaussianMarbles~\cite{stearns2024dynamicmarbles} & 22.32 & 0.129 \\
MonoNeRF~\cite{tian2023mononerf}        & 25.62 & 0.106 & MoSca~\cite{lei2025mosca}              & \cellcolor{tabsecond}26.72 & 0.070 \\
4DGS~\cite{Wu_2024_4dgs}                & 21.45 & 0.199 & \textbf{Ours}                          & \cellcolor{tabfirst}26.87 & \cellcolor{tabsecond}0.068 \\ 
Casual-FVS~\cite{lee2024casualfvs}      & 24.57 & 0.081 & & & \\
\bottomrule
\end{tabular}
}
\vspace{1mm}
\caption{Novel-view synthesis results on the NVIDIA Dataset~\cite{yoon2020novel}.}
\vspace{-13mm}
\label{tab:nvidia}
\end{table}

\subsection{Evaluations on NVIDIA Dataset}
\begin{figure}
  \centering
  \includegraphics[width=1\linewidth]{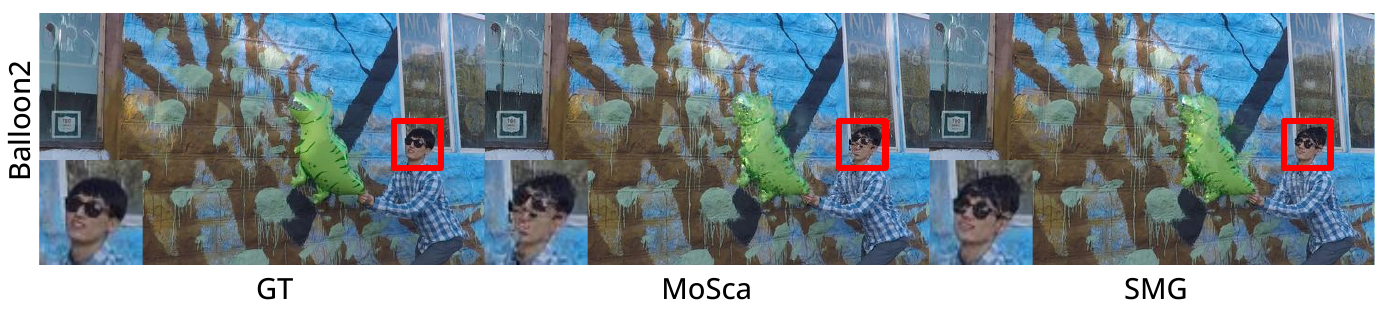}
  \vspace{-6mm}
  \caption{Qualitative results on the NVIDIA dataset.}
  \vspace{-3mm}
  \label{fig:nvidia}
\end{figure}

We conduct experiments on the NVIDIA dataset for novel-view synthesis following MoSca's~\cite{lei2025mosca} protocol (Table~\ref{tab:nvidia}). NVIDIA dataset is a real-world video dataset captured by front-facing cameras. 
Due to the narrow-baseline, the viewpoints of the training and testing frames are very similar. 
Most motions are fully observed and temporally stable, leaving few under-constrained regions where semantic local motion guidance plays a significant role. 
Nevertheless, SMG achieves comparable results to state-of-the-art methods.
SMG also achieves better motion propagation and preserves cleaner geometry. As shown in Fig.~\ref{fig:nvidia} (the human face in "Balloon2"), the proposed semantic motion propagation method effectively prevents motion leakage and contributes to cleaner geometry even in well-constrained scenes.

\begin{figure}[t]
\centering

\begin{minipage}{0.4\linewidth}
\centering
\includegraphics[width=\linewidth]{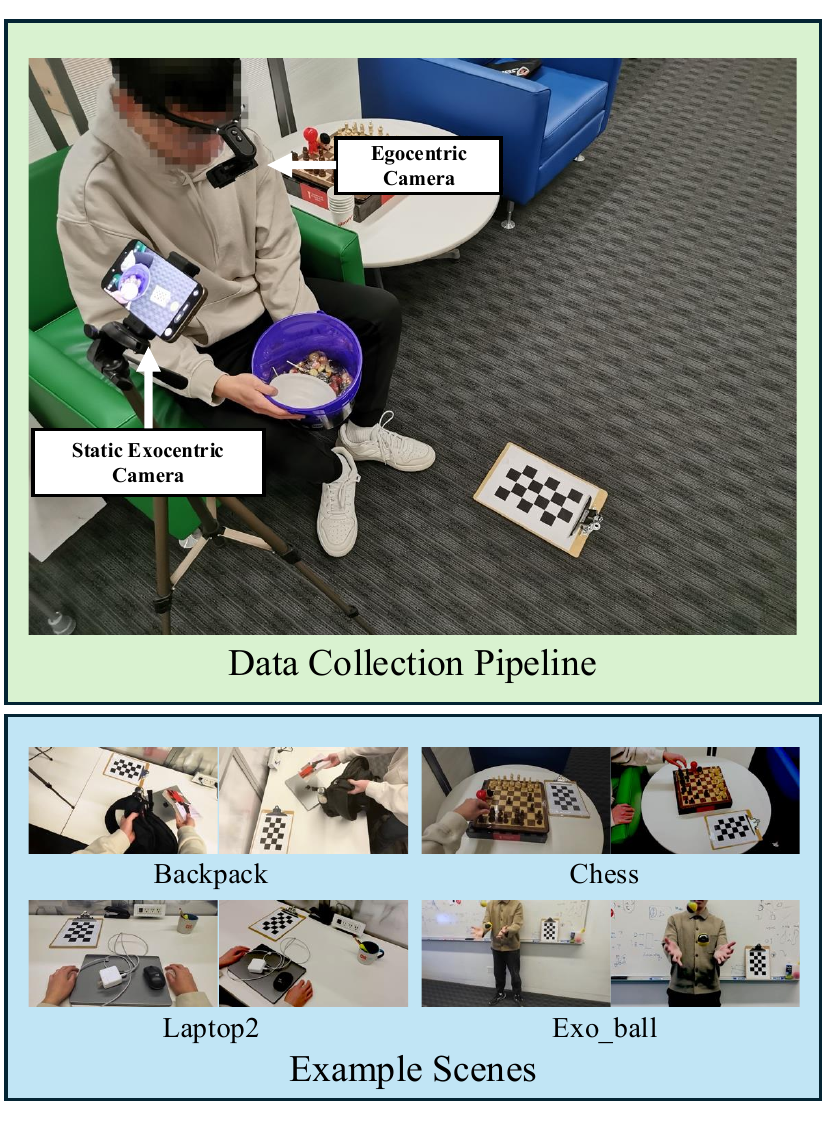}
\end{minipage}
\hfill
\begin{minipage}{0.58\linewidth}
\centering
\small
\begin{tabular}{@{}l|l|ccc@{}}
\toprule
Scene & Method & ~PSNR$\uparrow$~ & ~SSIM$\uparrow$~ & ~LPIPS$\downarrow$~ \\
\midrule
\multirow{3}{*}{backpack~}
 & MoSca~\cite{lei2025mosca} & 10.01 & 0.532 & 0.466 \\
 & OriGS~\cite{wu2025origs} & 10.22 & 0.529 & 0.479 \\
 & SMG & \bf{10.38} & \bf{0.542} & \bf{0.454} \\
\midrule
\multirow{3}{*}{chess}
 & MoSca~\cite{lei2025mosca} & 12.43 & 0.410 & 0.439 \\
 & OriGS~\cite{wu2025origs} & 12.89 & 0.412 & 0.460 \\
 & SMG & \bf{13.02} & \bf{0.445} & \bf{0.410} \\
\midrule
\multirow{3}{*}{laptop2}
 & MoSca~\cite{lei2025mosca} & 13.69 & 0.576 & 0.409 \\
 & OriGS~\cite{wu2025origs} & 13.85 & 0.570 & 0.430 \\
 & SMG & \bf{14.27} & \bf{0.581} & \bf{0.374} \\
\midrule
\multirow{3}{*}{exo\_ball}
 & MoSca~\cite{lei2025mosca} & 17.10 & 0.652 & 0.366 \\
 & OriGS~\cite{wu2025origs} & 17.13 & 0.660 & 0.357 \\
 & SMG & \bf{17.27} & \bf{0.668} & \bf{0.348} \\
\midrule
\multirow{3}{*}{All}
 & MoSca~\cite{lei2025mosca} & 13.31 & 0.542 & 0.421 \\
 & OriGS~\cite{wu2025origs} & 13.52 & 0.543 & 0.432 \\
 & SMG & \bf{13.74} & \bf{0.559} & \bf{0.396} \\
\bottomrule
\end{tabular}
\end{minipage}

\vspace{-2mm}
\caption{
Left: Data collection pipeline and example scenes. Right: Quantitative comparison on the proposed SMG dataset.
}
\vspace{-5mm}
\label{fig:dataset_table}
\end{figure}

\begin{figure}[t]
\centering
\includegraphics[width=\linewidth]{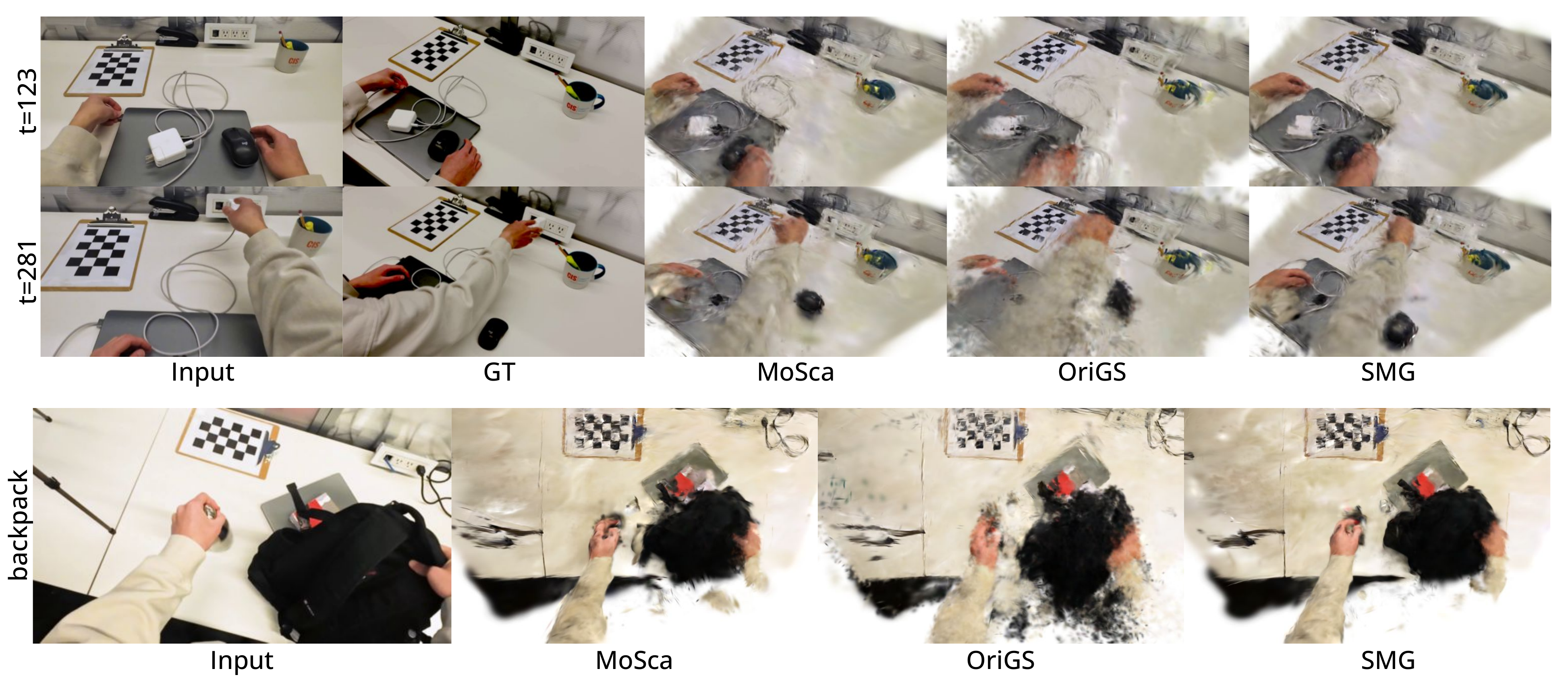}
\vspace{-3mm}
\caption{
Top: Qualitative comparison on the testing camera of the "laptop2" scene. Bottom: Novel-view visualization near the input cameras of the “Backpack” scene. 
}
\vspace{-6mm}
\label{fig:sand_bench_compare}
\end{figure}

\subsection{Evaluations on SMG Dataset}
While Dycheck provides multi-view videos with diverse real-world motion, most dynamics are dominated by simple object rotations (e.g., rotating humans, apples, or steering wheels), and camera motion is relatively mild. As a result, off-the-shelf priors such as depth and tracks can typically be estimated reliably.

To evaluate the robustness of 3D lifting field methods under more challenging conditions, we introduce the \textbf{SMG Dataset}, a real-world multi-view video dataset featuring complex camera motion and human–object interactions. Unlike Dycheck, which relies on multiple casually placed cameras, our setup pairs a moving egocentric camera with a static exocentric camera (Fig.~\ref{fig:dataset_table} (left)). The strong egocentric motion and intricate interactions produce substantially harder geometric conditions. We estimate per-frame camera poses using a PnP solver with calibrated 2D–3D correspondences, and provide metric depth maps predicted by the state-of-the-art metric depth estimator Depth Anything 3 (DA3)~\cite{lin2025da3}. Since egocentric depth predictions are often noisy, we prepare 4 scenes with post-processed DA3 depth for both quantitative and qualitative evaluation, while the remaining 16 scenes are used for qualitative evaluation.

We report the quantitative comparison between state-of-the-art methods~\cite{lei2025mosca, wu2025origs} and SMG in Fig.~\ref{fig:dataset_table} (right). SMG consistently outperforms MoSca and OriGS across all four scenes. 
The performance gap is smaller in the ``exo\_ball'' scene, where the camera baseline is small and the egocentric camera remains largely front-facing; under these conditions, both the observations and off-the-shelf priors are relatively reliable. In contrast, for scenes with large viewpoint changes and heavy occlusions, SMG outperforms MoSca and OriGS by a larger margin.

Qualitative results are shown in Fig.~\ref{fig:sand_bench_compare}. 
Due to the noisy predicted depth and tracks, novel-view synthesis becomes extremely challenging for dynamic Gaussian methods. We show qualitative comparisons on the test view of the "Laptop2" scene (top) and on a nearby view of the input camera of the "Backpack" scene (bottom). 
We highlight two representative failure cases of the previous methods. First, note the geometry and position of the mouse and the left hand over time, from frame 123 to 281 of "Laptop2". Previous methods incorrectly propagate the motion of the right arm to the static mouse, which is not visible in the training view, leading to clear drift in the rendered novel views.
In contrast, SMG restricts motion propagation within local semantic groups, preventing motion leakage across unrelated objects and preserving accurate mouse geometry and position.
The left hand in "Laptop2" and the laptop and backpack in "Backpack" illustrate a second failure mode.
Without explicitly modeling uncertainty during optimization, the previous methods become under-constrained when observations are limited, causing the object geometry to collapse. With the proposed C-ARAP and LRM, SMG maintains the correct object geometry and position even under limited observations and recovers plausible geometry once observations become available.

\begin{table}
  \centering
  \vspace{-5mm}
  \small
  \begin{tabular}{@{}l|ccc@{}}
    \toprule
     Method & ~EPE $\downarrow$~ & ~$\delta^{3D}_{.05} \uparrow$~ & ~$\delta^{3D}_{.10} \uparrow$~ \\
    \midrule
    MoSca~\cite{lei2025mosca} & 0.055 & 73.1 & 89.6 \\
    OriGS~\cite{wu2025origs} & 0.057 & 71.9 & 89.7 \\
    SMG & \bf{0.052} & \bf{74.1} & \bf{91.6} \\
    \bottomrule
  \end{tabular}
  \vspace{1mm}
  \caption{3D tracking results on Dycheck dataset.}
  \vspace{-15mm}
  \label{tab:motion}
\end{table}

\begin{figure}
  \centering
  \includegraphics[width=1\linewidth]{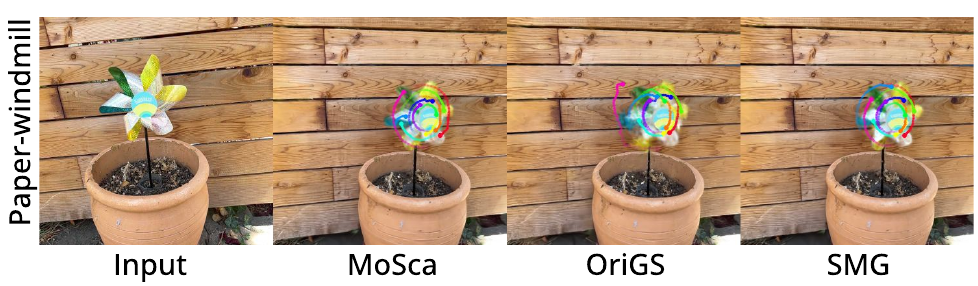}
  \vspace{-4mm}
  \caption{Qualitative results of 3D tracking on Dycheck dataset. (Zoom in to see the blue, purple and pink tracks.)}
  \vspace{-5mm}
  \label{fig:3dtrack}
\end{figure}

\subsection{Evaluations on 3D Tracking}
Since SMG is initialized with the lifted 2D tracks, it naturally supports long-range 3D point tracking. 
We further evaluate this capability on the five scenes (Apple, Block, Spin, Teddy, and Paper-windmill) of the Dycheck dataset. We report the 3D end-point-error (EPE) and the percentage of points that fall within a given threshold of the ground truth 3D location $\delta^{3D}_{.05}$ $\delta^{3D}_{.10}$, where the thresholds are 5cm and 10cm. 
SMG successfully models these local deformations, resulting in more accurate and semantically consistent 3D motion tracks. As shown in Table~\ref{tab:motion}, SMG outperforms MoSca and OriGS on all metrics.
We further show the qualitative results on the "Paper-windmill" scene in the Dycheck dataset (Fig.~\ref{fig:3dtrack}). With the proposed C-ARAP and LRM, the semantic motion graph propagates correct motion across the frames. 



\subsection{Ablation Study}

We conduct ablations to evaluate the impact of each component in our pipeline. The quantitative experiments (Table~\ref{tab:ablation}) are conducted on 4 Dycheck scenes (Apple, Spin, Space-out, Wheel). ``Base'' denotes a vanilla dynamic Gaussian Splatting~\cite{Wu_2024_4dgs}. ``SMG-only'' denotes the vanilla SMG without C-ARAP and LRM.

\begin{figure}
  \centering
    \vspace{-15mm}
  \includegraphics[width=1\linewidth]{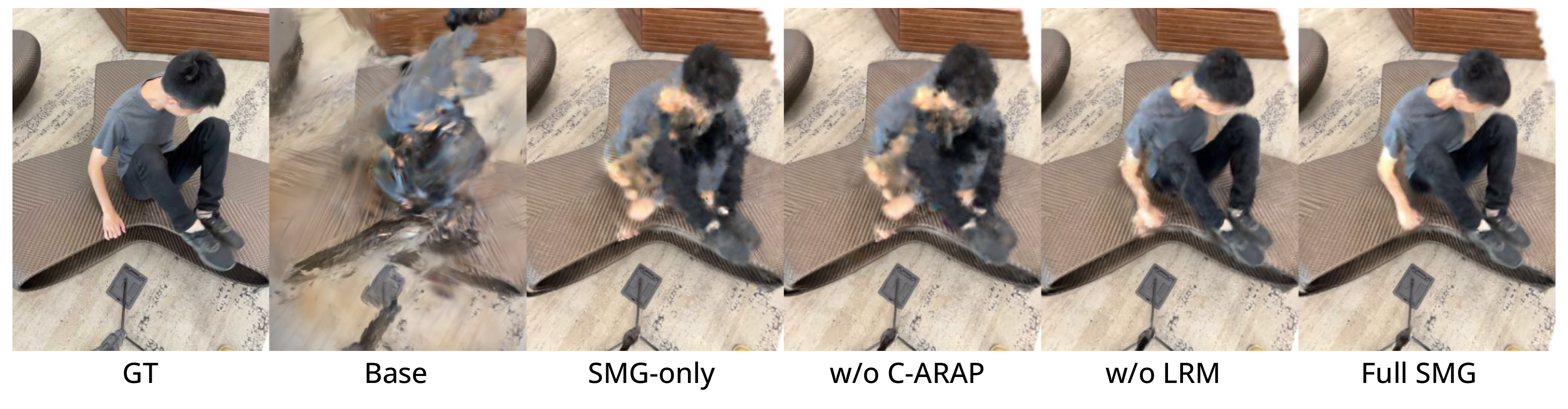}
  \vspace{-8mm}
  \caption{Qualitative ablation on the effectiveness of each component in the pipeline.}
  \vspace{-10mm}
  \label{fig:ablation quali}
\end{figure}

Compared to vanilla dynamic Gaussian Splatting, SMG-only already improves novel-view synthesis by regularizing Gaussian deformation with semantic motion structure. This is also reflected in qualitative ablation in Fig.~\ref{fig:ablation quali}. Without motion regularization, the vanilla 4DGS collapses under novel-view rendering, whereas SMG produces more plausible dynamic reconstructions. However, SMG alone does not reliably preserve topology during motion propagation, which limits its performance.
Adding C-ARAP further improves the results. As shown in Table~\ref{tab:ablation}, removing C-ARAP leads to a clear performance drop. The qualitative results explain this degradation. Without C-ARAP, the motion graph is more likely to suffer from topology drift, resulting in unstable Gaussian motion and degraded reconstructions. By enforcing locally consistent motion propagation under uncertainty, C-ARAP substantially improves novel-view stability.
LRM alone provides only limited improvement when directly applied to SMG. This is because LRM cannot correct topology drift by itself introduced by the uncertainty during optimization. Instead, LRM is most effective when combined with C-ARAP. 
Without LRM, C-ARAP may still be affected by the optimization uncertainty during the optimization, which can occasionally produce floating Gaussian artifacts, as shown in the right-hand region of Fig~\ref{fig:ablation quali}.

The full model combines the semantic motion graph with both uncertainty modeling components C-ARAP and LRM. It achieves the best quantitative performance and the most consistent qualitative reconstructions across frames, demonstrating the complementary benefits of semantic motion modeling and uncertainty-aware regularization.

\begin{table}
  \centering
  \vspace{-4mm}
  \small
  \begin{tabular}{@{}l|ccc@{}}
    \toprule
     Method & ~PSNR $\uparrow$~ & ~SSIM $\uparrow$~ & ~LPIPS $\downarrow$~ \\
    \midrule
    Base~\cite{Wu_2024_4dgs} & 12.92 & 0.451 & 0.598  \\
    SMG-only & 19.07 & 0.700  &  0.280  \\
    w/o C-ARAP~ & 19.09  & 0.703  & 0.281 \\
    w/o LRM  & 19.96 & 0.731 & 0.248  \\
    Full SMG & \bf{20.11} & \bf{0.738} & \bf{0.242} \\
    \bottomrule
  \end{tabular}
  \vspace{1mm}
  \caption{Ablations for evaluating the contribution of each pipeline component.}
  \label{tab:ablation}
\end{table}

\section{Discussion}
\label{sec:conclusion}

\noindent \textbf{Limitations.} Though SMG achieves state-of-the-art performance on novel-view synthesis benchmarks, a few challenges remain unresolved. Our method relies on off-the-shelf priors for initialization. Inaccuracies in these components may affect the performance of our model. SMG naturally improves as these upstream predictions become more accurate. Our method relies on per-scene optimization, which limits its scalability to large-scale or real-time applications. Recent advances~\cite{xu20254dgt} for feed-forward 4D Gaussian Splatting offer a promising direction toward more efficient and flexible dynamic reconstruction.

In summary, we present SMG, a novel monocular dynamic Gaussian framework that models Gaussian motion as structured semantic motion. 
To address uncertainty during Gaussian optimization, we introduce dynamic uncertainty modeling components, C-ARAP and LRM, which regularize motion in unconstrained regions using reliable graph nodes and preserve the dynamic object geometry. 
We further provide a multiview dataset with complex camera and scene motion that evaluates dynamic Gaussian Splatting under challenging real-world conditions. 
Our approach outperforms state-of-the-art methods for dynamic novel-view synthesis on real-world benchmarks and highlights the potential of semantic motion modeling for dynamic scene reconstruction.

\subsection*{Acknowledgments}

Research supported by the NVIDIA Academic Grant Program using 4 $\times$ RTX PRO 6000 Blackwell GPUs.



\clearpage  


%
%
\bibliographystyle{splncs04}
\bibliography{main}
\end{document}